# Leveraging Billions of Faces to Overcome Performance Barriers in Unconstrained Face Recognition

Yaniv Taigman  and  Lior Wolf
face.com
{yaniv, wolf}@face.com

**Abstract**

We employ the face recognition technology developed in house at face.com to a well accepted benchmark and show that without any tuning we are able to considerably surpass state of the art results. Much of the improvement is concentrated in the high-valued performance point of zero false positive matches, where the obtained recall rate almost doubles the best reported result to date. We discuss the various components and innovations of our system that enable this significant performance gap. These components include extensive utilization of an accurate 3D reconstructed shape model dealing with challenges arising from pose and illumination. In addition, discriminative models based on billions of faces are used in order to overcome aging and facial expression as well as low light and overexposure. Finally, we identify a challenging set of identification queries that might provide useful focus for future research.

## 1  Benchmark and results

The LFW benchmark [6] has become the de-facto standard testbed for unconstrained face recognition with over 100 citations in the face recognition literature since its debut 3 years ago. Extensive work [15, 14, 13, 5, 7, 4, 10, 3, 8, 9, 11, 16] has been invested in improving the recognition score which has been considerably increased since the first non-trivial result of 72% accuracy.

We employ face.com's r2011b[1] face recognition engine to the LFW benchmark without any dataset specific pre-tuning. The obtained mean accuracy is $91.3\% \pm 0.3$, achieved on the test set (view 2) under the unrestricted LFW protocol. Figure 1 (a) presents the ROC curve obtained in comparison to previous results. Remarkably, much of the obtained improvement is achieved at the conservative performance range, i.e., at low False Acceptance Rates (FAR).

---

[1]face.com has a public API service [1] which currently employs a previous version of the engine.

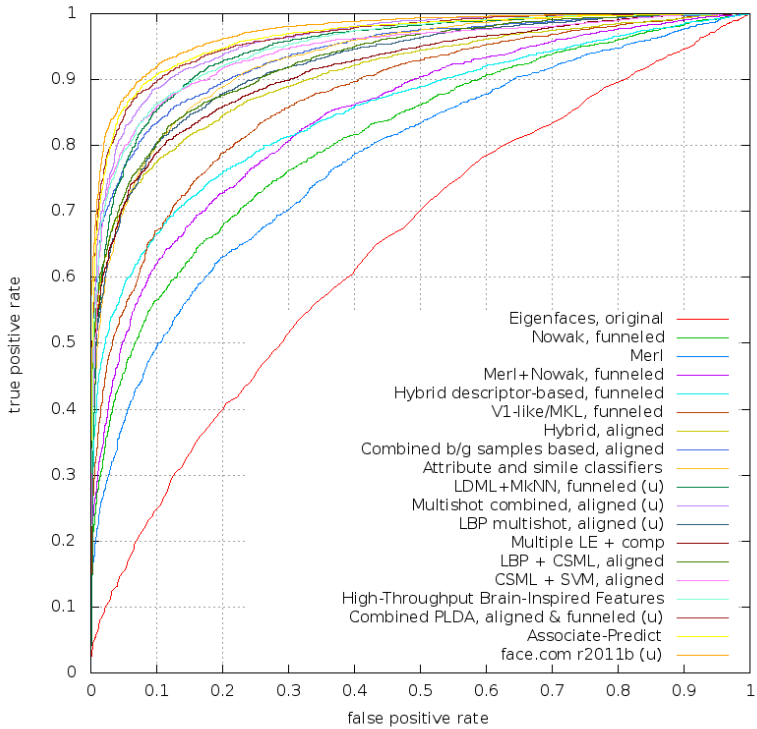 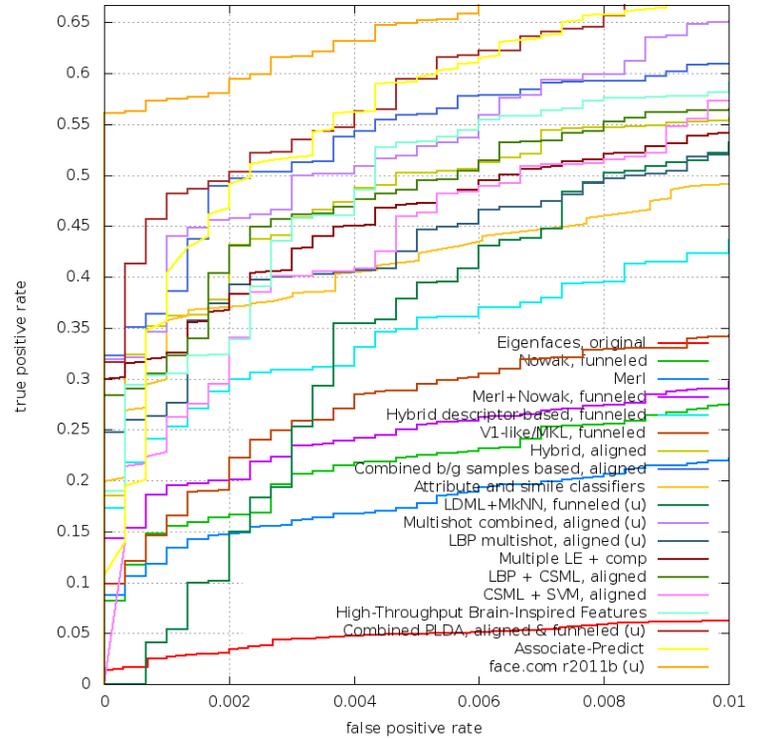

(a)                                                (b)

Figure 1: ROC curves for View 2 of the LFW data set. Each point on the curve represents the average over the 10 folds of (false positive rate, true positive rate) for a fixed threshold. (a) Full ROC curve. (b) A zoom-in onto the low false positive region. The proposed method is compared to scores currently reported in http://vis-www.cs.umass.edu/lfw/results.html

Specifically, for FAR=0 the recall (TPR) is over 55%, which is significantly higher than all previously reported results, as shown on Figure 1 (b).

As can be seen in Figure 6, the false matches arise in circumstances that are considerably difficult even for humans to recognize. This is often the result of extreme personal makeovers (much of LFW is concerned with celebrities) and challenging imaging conditions. Anecdotally, using the obtained results, the system was able to identify a newly discovered error among the thousands of labels of the benchmark when it discriminated clearly between the two basketball coaches named Jim O'Brien.

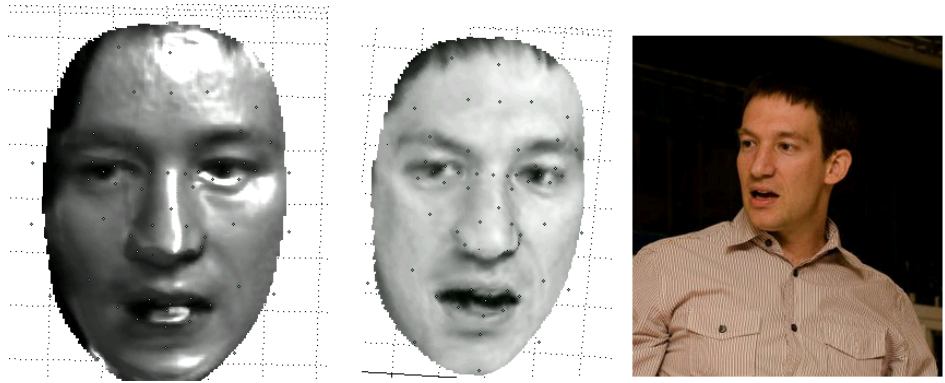

Figure 2: Pose correction (middle) to a non-frontal input 70x70 image (right) with arbitrary lightning (left)

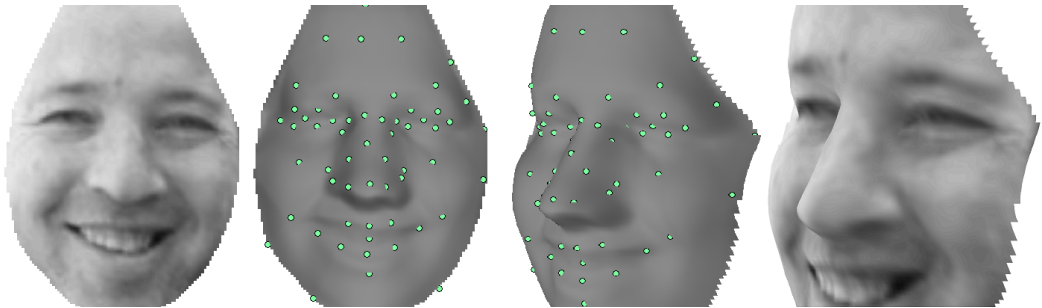

Figure 3: Input image (left most) and its 3D reconstructed model shown from different angels, with/without texture and anthropometric points

## 2 Methods

Face.com has been used by users and developers to index almost 31 billion face images of over 100,000,000 individuals. Leveraging this immense volume of data presents both a unique opportunity and an unusual challenge. The capability developed in house in order to make use of this data builds upon various achievements in scientific computation, database management and machine learning techniques. The run-time engine itself is a real-time one, able to process face detection together with recognition of over 30 frames per second on a single Intel 8-core server machine based on the Sandy Bridge architecture [12].

One key direction in which the large volume of data is utilized is in the development of a proprietary 3D face reconstruction engine. This engine is able to produce an accurate 3D model from a single unconstrained face image.

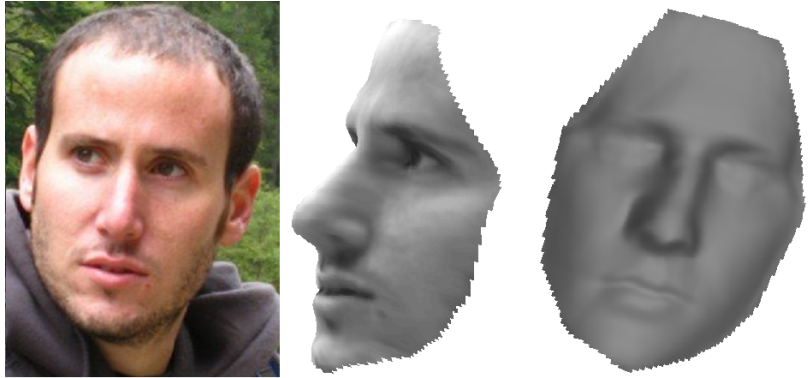

Figure 4: Another 3D reconstruction sample. Input image (left), Shape only rendering (right) with an arbitrary view rendering (middle)

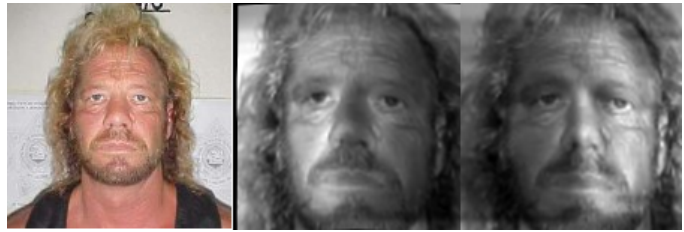

Figure 5: Duane Lee Chapman 0001 from the LFW dataset with arbitrary lightning imposed on its reconstructed 3D model

Unlike 3DMMs [2], face.com's 3D system works in real-time and is robust enough to handle general unconstrained imaging conditions in rather low-resolution images, see Figures 2, 3, and 4 for examples.

Once 3D reconstruction is obtained, two of the biggest challenges in face recognition become well defined and tractable. Namely, the face recognition engine is able to largely overcome pose and illumination variations. Pose is dealt by a normalization process in which all images are mapped to a frontal view. Unlike previous works [14, 13] [2] that tried to achieve view normalization without 3D modeling, outer plane rotation is accurately handled. The 3D model also enables the re-illuminating or rather delighting of the model once the parameters of the light sources are estimated, see Figure 5.

Some variations in face images of the same individual arise from aging or expression and are hard to model directly. By employing non-parametric discriminative models trained with tens of millions of data pieces, we are able to extract learned complex features that are invariant to these factors. Specifically, these features are based on building blocks that are selected exemplars from our repository, which are used to classify new probes as well as estimating attributes such as ethnicity, age and more.

---

[2]various other contributions have also employed LFW-a which is an aligned version of LFW obtained using the face.com API [1] as well.

Despite considerable improvement over state of the art results, performance is still not perfect, and some image pairs are mislabeled, see Figure 6. In order to promote the research of difficult cases, we are releasing[3] the full list of view 2's scores, i.e. 6000 similarity scores concatenated from the 10 splits, together with a subset list of these challenging pairs, that were misclassified by our system. Each mislabeled pair presents a rather unique challenge and therefore we estimate the risk of overfitting from studying these pairs as rather low. However, it is important to evaluate performance on these pairs only for systems that also achieve good performance in the official LFW benchmark.

---

[3]See: http://face.com/research

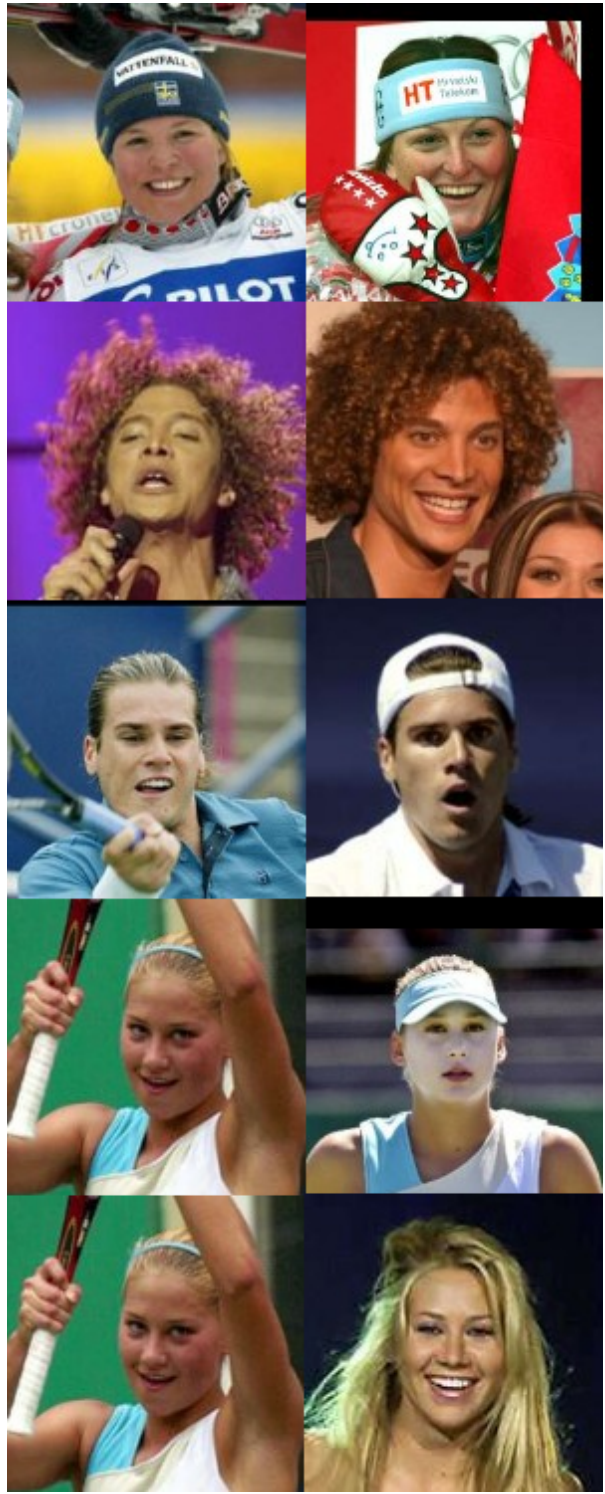

Figure 6: Examples of misclassified queries from the provided "Hard Pairs" split. In all five cases the correct label is "same"